# Classification of one family of 3R positioning Manipulators

Maher BAILI, Philippe WENGER, Damien CHABLAT
Institut de Recherche en Communications et Cybernétique de Nantes UMR 6597
1, rue de la Noë, BP 92101, 44312 Nantes Cedex 03 France
Maher.Baili@irccyn.ec-nantes.fr , Philippe.Wenger@irccyn.ec-nantes.fr , Damien.Chablat@irccyn.ec-nantes.fr

## ABSTRACT

The aim of this paper is to classify one family of 3R serial positioning manipulators. This categorization is based on the number of cusp points of quaternary, binary, generic and non-generic manipulators. It was found three subsets of manipulators with 0, 2 or 4 cusp points and one homotopy class for generic quaternary manipulators. This classification allows us to define the design parameters for which the manipulator is cuspidal or not, i.e., for which the manipulator can or cannot change posture without meeting a singularity, respectively.

## 1. INTRODUCTION

There is a strong relationship between global kinematic properties and the topology of singularities. Few authors are interested in global kinematic analysis of manipulators. Most of them work on control and trajectory planning or on manipulator workspace analysis. However, the geometry and topology of the singularities are a very important way for the analysis and classification of the kinematic properties of manipulators. In [1], we have a detailed analysis of 3R manipulator singularities. In [2], the notion of genericity was introduced. A manipulator is generic if its singularities are generic (they do not intersect in the joint space). Non-generic manipulators form hypersurfaces dividing the space of manipulators into different sets of generic ones. Consequently, most manipulators are generic. In [3], a categorization of all quaternary generic 3R positioning manipulators was done using homotopy classes. It was found exactly eight classes of homotopic quaternary generic 3R manipulators. The goal of this paper is to classify one family of positioning serial 3R manipulator. Moreover, an exhaustive classification of non-generic and binary manipulators is given. This study serves as an efficient tool for the categorization of cuspidal and non-cuspidal manipulators. This paper is organized as follow. Section 2 recalls some notions like singularities, cuspidality, genericity and homotopy class. Section 3 describes the family of 3R serial manipulator, which will be used all along this paper to make the categorization. In section 4, we analyse the results found. Some examples are given in section 5. Finally, last section concludes this paper.

## 2. PRELIMINARIES

### 2.1 SINGULARITIES

This paper deals with serial 3R positioning manipulators and only positioning singularities (referred to as "singularity" in the rest of the paper) are considered. A singularity can be characterized by a set of joint configurations that nullifies the determinant of the Jacobian matrix. They divide the joint space into at least two domains called aspects [4]. The aspects are the maximal free-singularity domains in the joint space.

[1] defines the critical point surfaces as the connected and continuous subset of singularities. Their corresponding images in the workspace are defined as critical value surfaces. The critical value surfaces divide the workspace into different regions with different number of inverse kinematic solutions or postures [5].

For a 3R manipulator, the joint space has the structure of a 3-dimentionnal torus, which can be reduced to a 2-dimentionnal-torus ($\theta_2, \theta_3$) because the manipulators do not depend on $\theta_1$. In order to clarify Figure 2, the torus is cut along its generators, so the singularities are plotted in a $2\pi$ dimensional space. We must identify the opposite side of the square to keep the topology of the torus.

### 2.2 CUSPIDAL MANIPULATORS

A cuspidal manipulator is one that can change posture without meeting a singularity. The existence of such manipulators was discovered simultaneously by [1] and [6]. In [7], a theory and methodology were introduced to characterize new uniqueness domains in the joint space of cuspidal manipulators. Some examples of these manipulators are studied and analysed in [8]. The only possible region of the workspace where a cuspidal manipulator can change posture without meeting singularity, is a region with four inverse kinematic solutions. Characterization of cuspidal manipulators has been a serious difficulty. Obviously, observation of several examples of manipulators gave rise to some conjectures by authors. In fact, some of them think that manipulators with simplifying geometric conditions like intersecting, orthogonal or parallel joint axes cannot avoid singularities when changing posture [5-12]. Others think that manipulators with arbitrary kinematic parameters are cuspidal [1-13]. Neither the first nor the second idea can be stated in a general way. In [9], a new characterization of cuspidal manipulators was done: a 3-DOF positioning manipulator can change posture without meeting a singularity if and only if there exists at least one point in its workspace with exactly three coincident inverse kinematic solutions and such a point is called a cusp point. Figure 1 shows in a cross-section of the workspace, the critical value surfaces for one cuspidal manipulator (D-H parameters modified [10]: $d_2 = 1$, $d_3 = 2$, $d_4 = 1.5$, $r_2 = 1$, $r_3 = 0$, $\alpha_2 = -90deg$ and $\alpha_3 = 90deg$). There are four cusp points and two regions with four and two possible postures, respectively.

Numeric and graphic methods are used to check the existence conditions of a cusp point. Consequently, it provides a useful tool for the purpose of manipulator design. In general cases, it is not possible to write the existence conditions of cusp points in an explicit, amenable expression of the DH-parameters [11].

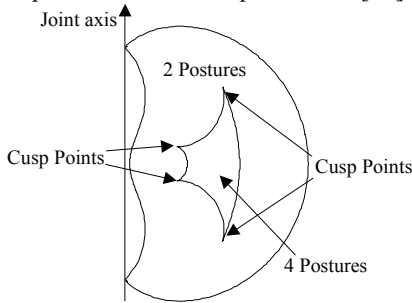

*Figure 1: Cusp points in the workspace section of a cuspidal manipulator.*

### 2.3 GENERIC MANIPULATOR

In [2], a generic manipulator is defined as one having no intersection of its smooth singularity surfaces in the joint space. Furthermore, a generic 3R manipulator must satisfy the following two conditions:

(1) The jacobian matrix has rank 2 at all critical points.

(2) All singular points, $\theta_s$, must satisfy the following condition:

$$\frac{\partial [\det(\mathbf{J}(\theta_s))]}{\partial \theta_i} \neq 0 \text{ for at least one, } i=1,2 \quad (1)$$

Simplifications in manipulator geometry (like intersecting or parallel joint axis) often lead to non-genericity and most of industrial manipulators are non-generic. However, many non-generic manipulators have complicated DH-parameters [12-13]. Generic manipulators have stable global kinematic properties under small changes in their design parameters.

### 2.4 HOMOTOPY CLASSES

Homotopy classes were defined in [3] only for generic quaternary manipulators. A quaternary manipulator is defined as one having 4 inverse kinematic solutions. A binary manipulator has only 2 solutions. Two quaternary generic manipulators are homotopic if the singularity surfaces of one manipulator can be smoothly deformed to the singularity surfaces of the other. [13] shows that two homotopic manipulators have the same multiplicity of their kinematic maps. This result can be used to say that homotopic manipulators have the same maximum number of inverse kinematic solution by aspect. Thus, if one manipulator is cuspidal (resp. non-cuspidal), all manipulators homotopic to it are cuspidal (resp. non-cuspidal).

One singularity surface forms a loop on the surface of the torus $(\theta_2, \theta_3)$. Consequently, there are as many homotopy classes as ways of encircling the two generators of the torus. Figure 2 shows three different homotopy classes. Lines L1 and L2 represented in the square ($-\pi \leq \theta_2 \leq \pi$, $-\pi \leq \theta_3 \leq \pi$) correspond to circles along the $\theta_2$-generator and the $\theta_3$-generator of the torus respectively. However, L3 does not encircle any of the two generators, so it is homotopic to one point.

The homotopy class of a manipulator is denoted $n$ $(n_2, n_3)$, where:

- $n$: number of singularity surfaces in joint space.
- $n_2$: number of time the singularity surface encircles the $\theta_2$-generator.
- $n_3$: number of time the singularity surface encircles the $\theta_3$-generator.

To identify the homotopy class of one manipulator, the idea is to take one singularity surface, to count le number of "jumps" between two opposite sides of the square representation. At each jump, $n_2$ and $n_3$ are increased or decreased according to whether a jump occurs from $-\pi$ to $\pi$ or from $\pi$ to $-\pi$.

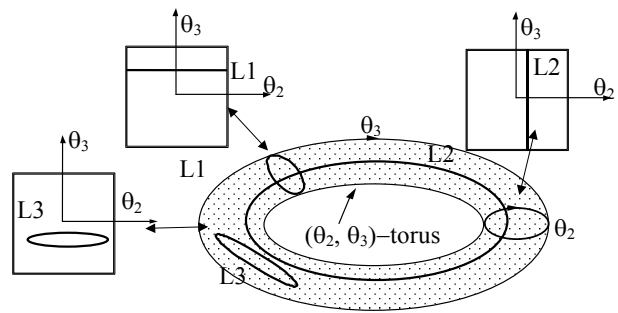

*Figure 2: Some loops of homotopy classes on the torus*

For example, the homotopy class of L1 (resp. L2, L3) is (1,0), (resp. (0,1), (0,0)).

The number and the homotopy class of the singularity surfaces define a set of homotopic generic manipulators.

The set of all 3R positioning manipulators is divided into subsets of homotopic generic manipulators separated by subsets of non-generic manipulators [3].

## 3. PROBLEM FORMULATION

The kinematic structures of most industrial manipulators are frequently decoupled into positioning and orientation devices. Their positioning structures are generally such that $\alpha_2 = \pm 90$ deg, $\alpha_3 = 0$ deg, $r_2=0$ or $d_2=0$ and $r_3=0$ (like the "PUMA" manipulator, all the manipulators such that $\alpha_3 = 0$ are non-generic, quaternary and non-cuspidal [3-12]). In this paper, however, we study a family of positioning manipulators such that $\alpha_2 = \pm 90$ deg, $\alpha_3 = \pm 90$ deg and $r_3=0$ (Figure 3). This is a more interesting family because such manipulators can be binary or quaternary, generic or non-generic, cuspidal or non-cuspidal. We normalise $d_3$, $d_4$ and $r_2$ by $d_2$. So, we have only 3 parameters to consider: $d_3/d_2$, $d_4/d_2$ and $r_2/d_2$.

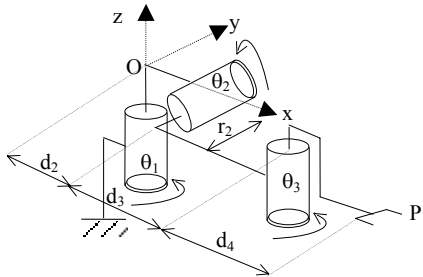

*Figure 3: 3R manipulators studied.*

The direct kinematic equations are defined by:

$$\begin{cases} x = (d_3 + d_4 \cos\theta_3)(\cos\theta_1 \cos\theta_2) \\ \quad -(r_2 + d_4 \sin\theta_3)\sin\theta_1 + \cos\theta_1 \\ y = (d_3 + d_4 \cos\theta_3)(\sin\theta_1 \cos\theta_2) \\ \quad +(r_2 + d_4 \sin\theta_3)\cos\theta_1 + \sin\theta_1 \\ z = -(d_3 + d_4 \cos\theta_3)\sin\theta_2 \end{cases} \quad (2)$$

Our goal is to express a condition to know if the manipulator is cuspidal or not. Equations (2) can be written in $\theta_3$ only [14]. We obtain the equation below:

$$m_5 \cos\theta_3^2 + m_4 \sin\theta_3^2 + m_3 \cos\theta_3 \sin\theta_3 \\ + m_2 \cos\theta_3 + m_2 \sin\theta_3 + m_0 = 0 \quad (3)$$

Where:

$$\begin{cases} m_0 = -x^2 - y^2 + r_2^2 + \dfrac{(R+1-L)^2}{4} \\ m_1 = 2r_2 d_4 + (L-R-1)d_4 r_2 \\ m_2 = (L-R-1)d_4 d_3 \end{cases} \text{ and } \begin{cases} m_3 = 2r_2 d_3 d_4 \\ m_4 = d_4^2(r_2^2+1) \\ m_5 = d_3^2 d_4^2 \end{cases}$$

With: $R = x^2 + y^2 + z^2$ and $L = d_4^2 + d_3^2 + r_2^2$

We obtain the polynomial in $t = \tan(\theta_3/2)$:

$$P(t) = at^4 + bt^3 + ct^2 + dt + e \quad (4)$$

With:

$$\begin{cases} a = m_5 - m_2 + m_0 \\ b = -2m_3 + 2m_1 \\ c = -2m_5 + 4m_4 + 2m_0 \end{cases} \text{ and } \begin{cases} d = 2m_3 + 2m_1 \\ e = m_5 + m_2 + m_0 \end{cases}$$

To say if the manipulator is cuspidal or not, the polynomial (4) (of degree 4 in $t$ and with coefficients function of $x, y, z, d_3, d_4, r_2$) must admit real triple roots. This is equivalent to solve the system:

$$\begin{cases} P(t, d_3, d_4, r_2, R, z) = 0 \\ \dfrac{\partial P}{\partial t}(t, d_3, d_4, r_2, R, z) = 0 \\ \dfrac{\partial^2 P}{\partial t^2}(t, d_3, d_4, r_2, R, z) = 0 \\ \dfrac{\partial^3 P}{\partial t^3}(t, d_3, d_4, r_2, R, z) \neq 0 \end{cases}$$

Where:

- $t, R, z$ are the variables.
- $d_3, d_4, r_2$ are the parameters.

We note that one solution in $t$ lifts unique 3-uplet $(\theta_1, \theta_2, \theta_3)$ except when $z = 0$, that can be treated like a particular case.

[15] computed a partition of the space of parameters, in which the number of real solutions is constant in each cell. This permits us to select one manipulator in each cell that can be considered like a representative manipulator of this cell. Under the hypotheses $Z = z^2 > 0$ and $R - Z > 0$, elimination of the variables $t, Z, R$ after many operations and using Groebner Basis allows us to find one triangular system. Its regular roots represent solutions of the problem. Analysis of this system leads to only two equation surfaces dividing the parameter space into different zones [15].

These surfaces equations are defined by:

$d_3^2 - d_4^2 + r_2^2 = 0$ and

$$d_4^2 d_3^6 - d_4^4 d_3^4 + 3d_4^2 d_3^4 r_2^2 - 2d_4^2 d_3^4 + 2d_4^4 d_3^2$$
$$-2d_4^4 d_3^2 r_2^2 + d_4^2 d_3^2 + 3d_4^2 d_3^2 r_2^4 - d_3^2 r_2^2 - 2d_4^4 r_2^2$$
$$-d_4^4 + d_4^2 r_2^6 + d_4^2 r_2^2 + 2d_4^2 r_2^4 = 0$$

CAD (Cylindrical Algebraic Decomposition) adapted to such polynomial makes representation of possible cells. Our work is limited only to rational test points because we are interested in cells of maximal dimension (no computation with real algebraic numbers).

Finally, we obtain one test point by cell, with positive rational coordinates and a set of hypersurfaces containing the other cells of the CAD. In practice, 105 maximal dimension cells are provided. Figure 4 depicts a CAD projection in the plane ($d_3$, $r_2$) of the partition of parameter space ($d_3$, $r_2$, $d_4$).

Here, we can see 5 zones in plane ($d_3$, $r_2$) and over everyone (in direction of $d_4$), we have 7 cells. For example, we have 35 cells in zone1 and 28 in zone2.

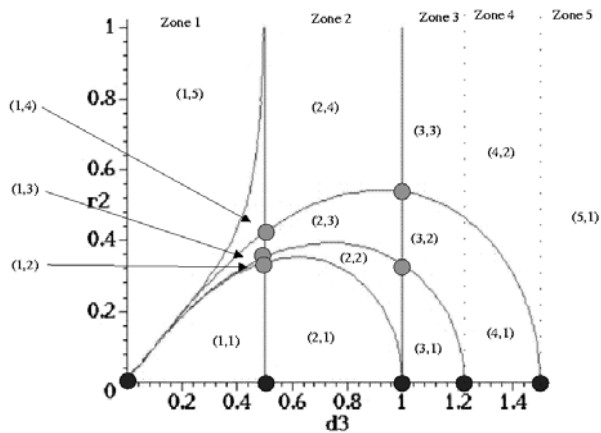

*Figure 4: projection of the partition of parameters' space.*

## 4. ANALYSIS WITH NUMBER OF CUSP POINTS

As it was said in the previous paragraph, the idea is to take one test point in each cell and to draw its workspace in order to know whether it is cuspidal or not. If the corresponding manipulator is cuspidal, it has the same number of cusp points and kinematic properties as all other manipulators in the cell.

We distinct two kinds of manipulators:

- Manipulator having four solutions for the inverse kinematic problem (quaternary manipulator).
- Manipulator having two solutions for the inverse kinematic problem (binary manipulator).

Classification into homotopy classes, correspond only for quaternary manipulators.

There exist three subsets of manipulators with 0, 2 and 4 cusp points, respectively. Table 1 shows that all non-cuspidal manipulators (0 cusp point) are binary and generic. In addition, these manipulators may have 2 or 4 aspects. Table 2 shows that all manipulators having two cusp points are quaternary, non-generic and have 4 aspects. On the other hand, Table 3 shows that manipulators with four cusp points are all quaternary and either generic with homotopy class 2(1, 0) and with 2 aspects, or quaternary non-generic with 4 aspects.

Only one homotopy class 2(1, 0) is found in this work among the eight possible homotopy classes found in [3] which are: {(1(0,0), 2(0,0), 1(0,0)+2(1,0), 2(1,0), 4(1,0), 2(0,1), 2(1,1), 2(2,1)}. The two cells (1,3,4) and (4,1,4) and all cells of Table 2 contain non-generic manipulators.

| Binary generic manipulators with 2 aspects | Binary generic manipulators with 4 aspects |
|---|---|
| (1,1,1); (1,1,2); (1,2,1); (1,2,7); (1,3,1); (1,4,1); (1,5,1). | (1,1,6);(1,1,7); (1,2,6); (1,3,6); (1,3,7); (1,4,6); (1,4,7); (1,5,5); (1,5,6); (1,5,7). |
| (2,1,1); (2,1,2); (2,2,1); (2,3,1); (2,4,1). | (2,1,7); (2,2,7); (2,3,7); (2,4,7). |
| (3,1,1); (3,2,1); (3,3,1). | |
| (4,1,1); (4,2,1). | |
| (5,1,1). | |

*Table 1: Cells with 0 cusp point*

| Quaternary non-generic manipulators with 4 aspects |
|---|
| (1,1,5); (1,2,5); (1,3,5); (1,4,4); (1,4,5); (1,5,4). |
| (2,1,5); (2,1,6); (2,2,5); (2,2,6); (2,3,5); (2,3,6); (2,4,4); (2,4,5); (2,4,6). |
| (3,1,5); (3,1,6); (3,1,7); (3,2,5); (3,2,6); (3,2,7); (3,3,4); (3,3,5); (3,3,6); (3,3,7). |
| (4,1,5); (4,1,6); (4,1,7); (4,2,4); (4,2,5); (4,2,6); (4,2,7). |
| (5,1,4); (5,1,5); (5,1,6); (5,1,7). |

*Table 2: Cells with 2 cusp points*

| *Quaternary generic with Homotopy class 2(1,0) and 2 aspects* | *Quaternary non-generic with 4 aspects* |
|---|---|
| (1,1,3); (1,1,4); (1,2,2); (1,2,3); (1,2,4); (1,3,2); (1,3,3); (1,4,2); (1,4,3); (1,5,2); (1,5,3). | (1,3,4). |
| (2,1,3); (2,1,4); (2,2,2); (2,2,3); (2,2,4); (2,3,2); (2,3,3); (2,3,4); (2,4,2) (2,4,3). | |
| (3,1,2); (3,1,3); (3,1,4); (3,2,2); (3,2,3); (3,2,4); (3,3,2) (3,3,3). | |
| (4,1,2); (4,1,3); (4,2,2); (4,2,3). | (4,1,4). |
| (5,1,2); (5,1,3). | |

*Table 3: Cells with 4 cusp points*

## 5. EXAMPLES

In this section, some examples of manipulators are provided. Figure 5 depicts the singularity surfaces (resp. critical value surface in a cross section of workspace) in $\theta_2$, $\theta_3$ (resp. workspace section defined by plane $\rho = \sqrt{x^2 + y^2}$, $z$). The number of inverse kinematic solutions in each region of the workspace is given. Figure 5 shows all possible shapes of joint spaces and workspaces met in this classification. Table 4 illustrates some information about proprieties of manipulators like their homotopy class (only for generic quaternary ones) and their cell belonging.

| Fig. 5 | Cell number | Nb. Aspect | Cuspidal | Class | DHM-parameters | | |
|---|---|---|---|---|---|---|---|
| | | | | | $d_3$ | $r_2$ | $d_4$ |
| (a) | (1,1,1) | 2 | No | binary | 0.21 | 0.1 | 0.05 |
| (b) | (1,2,6) | 4 | No | binary | 0.21 | 0.19 | 0.25 |
| (c) | (1,3,4) | 4 | Yes | n.g* | 0.21 | 0.2 | 0.21 |
| (d) | (4,2,2) | 2 | Yes | 2(1,0) | 1.36 | 0.35 | 0.75 |
| (e) | (2,4,5) | 4 | Yes | n.g* | 0.75 | 0.52 | 0.85 |
| (f) | (3,1,6) | 4 | Yes | n.g* | 1.11 | 0.13 | 1.4 |
| (g) | (5,1,1) | 2 | No | binary | 1.97 | 1 | 0.1 |
| (h) | (5,1,3) | 2 | Yes | 2(1,0) | 1.97 | 1 | 1.54 |

*Table 4: Some examples of manipulators*

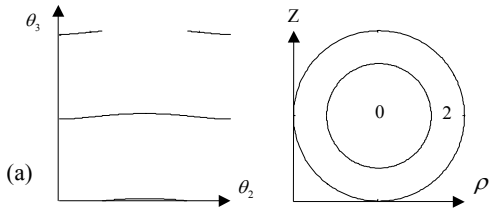

(a)

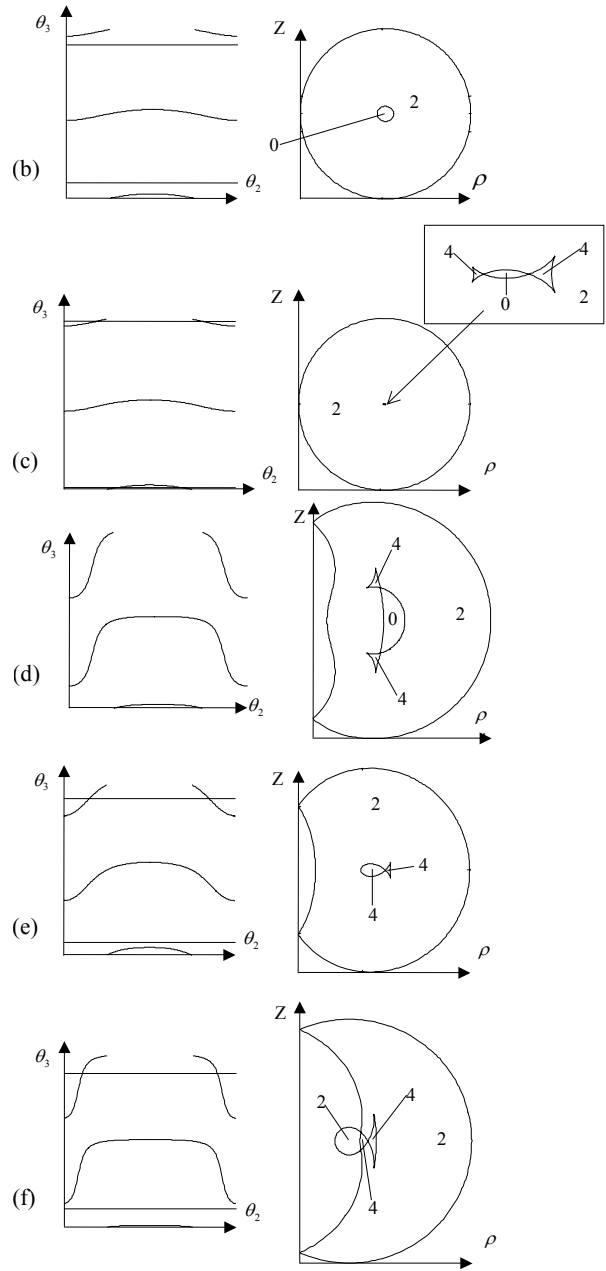

(b)

(c)

(d)

(e)

(f)

---

* *n.g* = non-generic.

Submitted to ICAR 2003

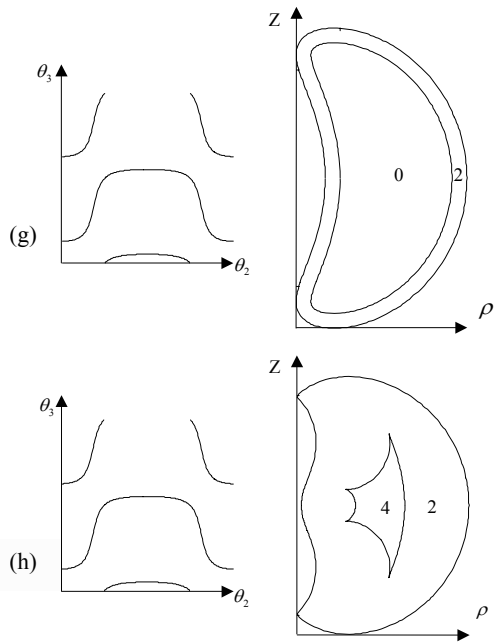

*Figure 5: Some examples of manipulators*

## 6. CONCLUSIONS

In this paper, we have provided a classification of one family of positioning manipulators such that $\alpha_2 = \pm 90$ *deg*, $\alpha_3 = \pm 90$ *deg* and $r_3=0$. This classification was based on the number of cusp points. Parameter space was divided into 105 cells. In each cell, all manipulators have either 0, 2 or 4 cusp points. In addition to that, a complete classification of all manipulators of the family studied was provided using different criteria such as number of aspects, generic/non-generic and binary/quaternary manipulators.

First, all binary manipulators are generic and non-cuspidal and can have 2 or 4 aspects. Second, all quaternary non-generic manipulators are cuspidal, have 4 aspects and can have 2 or 4 cusp points. Finally, quaternary generic manipulators have 2 aspects and belong to the homotopy class 2(1, 0): the only homotopy class founded among the eight possible.

Future research work is to include the complete classification of a larger family of manipulators by relaxing $r_3$ ($r_3 \neq 0$).

## 7. AKNOWLEDGMENTS

This research was partially supported by CNRS (Project MATHSTIC "Cuspidal Robots and Triple Roots).